\documentclass[a4paper,10pt,twocolumn]{article}

\usepackage[english]{babel}

\usepackage[a4paper,top=2cm,bottom=2cm,left=2cm,right=2cm,marginparwidth=1.75cm]{geometry}

\usepackage{nameref} 

\usepackage{newtxtext,newtxmath} 

\usepackage{graphicx} 
\usepackage{float}

\usepackage{multirow} 
\usepackage{array}

\usepackage{mathtools}
\newcolumntype{R}{>{\raggedright\arraybackslash}p{0.15\textwidth}}
\usepackage{newtxtext,newtxmath} 

\usepackage{xcolor}
\usepackage[colorlinks=true, allcolors=blue]{hyperref}

\usepackage{algorithm}
\usepackage{algpseudocode}

\usepackage{authblk}
\usepackage{mathrsfs}

\usepackage[title]{appendix}

\providecommand{\keywords}[1]
{
  \small	
  \textbf{\textit{Keywords---}} #1
}

\title{Deep convolutional forest: a dynamic deep ensemble approach for spam detection in text}

\newcommand{\orcid}[1]{\href{https://orcid.org/#1}{\includesvg[width=16pt]{orcidlogo}}}

\author[1]{Mai A. Shaaban\thanks{Corresponding Author: mai.shaaban@alexu.edu.eg}}
\author[2]{Yasser F. Hassan}
\author[3]{Shawkat K. Guirguis}

\affil[1]{Department of Mathematics and Computer Science, Faculty of Science, Alexandria University, Alexandria, Egypt}
\affil[2]{Faculty of Computers and Data Science, Alexandria University, Alexandria, Egypt}
\affil[3]{Institute of Graduate Studies and Research, Alexandria University, Alexandria, Egypt}

\date{}

\begin{document}

\twocolumn[
  \begin{@twocolumnfalse}
    \maketitle
    \begin{abstract}
    The increase in people's use of mobile messaging services has led to the spread of social engineering attacks like phishing, considering that spam text is one of the main factors in the dissemination of phishing attacks to steal sensitive data such as credit cards and passwords. In addition, rumors and incorrect medical information regarding the COVID-19 pandemic are widely shared on social media leading to people's fear and confusion. Thus, filtering spam content is vital to reduce risks and threats. Previous studies relied on machine learning and deep learning approaches for spam classification, but these approaches have two limitations. Machine learning models require manual feature engineering, whereas deep neural networks require a high computational cost. This paper introduces a dynamic deep ensemble model for spam detection that adjusts its complexity and extracts features automatically. The proposed model utilizes convolutional and pooling layers for feature extraction along with base classifiers such as random forests and extremely randomized trees for classifying texts into spam or legitimate ones. Moreover, the model employs ensemble learning procedures like boosting and bagging. As a result, the model achieved high precision, recall, f1-score and accuracy of 98.38\%.
    
    \vspace{0.2cm}
    \keywords{Ensemble methods, Deep learning, Machine learning, Spam classification, Text messages}
    
    \end{abstract}
  \end{@twocolumnfalse}
  \vspace{1cm}

]

\section*{Introduction}
\label{sec:intro}

Mobile messaging service has become one of the most common means of communication among people since they allow individuals to communicate with one another at any time and from any location. Besides, there is a vast number of messaging apps that provide their service for free. According to statistics, 95\% of mobile messages in the USA are read and responded to within three minutes of receiving \cite{Grossbard2021}. In addition, Short Message Service (SMS) offers businesses an enormous chance to communicate and interact with customers as 48\% of consumers prefer direct communication from businesses via SMS \cite{Grossbard2021}. As a result, users are prone to SMS attacks such as spam and phishing, especially users who lack awareness about cyber threats.

Spam text is any undesired text transmitted to people without their permission and may include a link to a phone number to call, a link to open a website, or a link to download a file. Thus, an attacker can masquerade as a trusted entity and exploit spam texts by attaching malicious links so that victims may be duped into clicking a harmful link, resulting in installing malware or revealing sensitive information, including login credentials and credit card numbers \cite{Goel2018,Jain2020}. For example, phishing attacks can occur by sending fake messages for users telling them to reset their passwords to be able to login to Facebook, Twitter, or any other platform \cite{Jain2021}. Besides, spammers can share misleading information about the COVID-19 pandemic causing a negative impact on society \cite{Rao2021}. Therefore, filtering spam texts is crucial to protect users against social engineering attacks, mobile malware and threats.

Previous studies in the area of spam classification focused on using machine learning algorithms \cite{Akinyelu2021}, but these algorithms require prior knowledge and domain expertise for identifying useful features in order to achieve accurate classification \cite{Roy2020}. Furthermore, researchers proposed deep learning approaches to detect spam \cite{Rao2021}. However, deep neural networks require much effort in tuning hyper-parameters \cite{Zhou2019}. Besides, they require massive data for training to predict new data accurately. Consequently, they require a high computational cost \cite{Zhou2019}.

To overcome the high complexity of deep learning models and to reduce the effort spent in tuning hyper-parameters, Zhou and Feng \cite{Zhou2019} developed multi-grained cascade forest (gcForest), a decision tree ensemble approach that can be applied to different classification tasks and has much fewer hyper-parameters than deep learning neural networks. Ensemble methods \cite{zhou2012ensemble} train multiple base models to produce a single best-fit predictive model. Kontschieder et al. \cite{Kontschieder2015} demonstrated that employing ensemble approaches like random forests \cite{Breiman2001} aided by deep learning model features can be more effective than solely using a deep neural network. gcForest applies multi-grained scanning for extracting features and employs a cascade structure (i.e., layer-by-layer processing of raw features) for learning. Inspired by gcForest, this paper enhances the procedure of feature engineering by replacing the multi-grained scanning with convolutional layers and pooling layers to capture high-level features from textual data. The motivation for using gcForest as a baseline for this paper is that gcForest is the first deep learning model that trains data without relying on neural networks and backpropagation, as the authors claimed {\cite{Zhou2019}}.

This paper introduces a dynamic (self-adaptive) deep ensemble mechanism to overcome the stated limitations of machine learning and deep learning approaches for detecting spam texts. The main contributions of this paper are as follows:
\begin{itemize}
\item[--] Implementing a dynamic deep ensemble model called Deep Convolutional Forest (DCF).
\item[--] Extracting features automatically by utilizing convolutional layers and pooling layers.
\item[--] Determining the model complexity automatically so that the model can perform accurately on both small-scale data and large-scale data.
\item[--] Classifying text into Spam or Ham (Not-Spam), achieving a remarkable accuracy.
\end{itemize}

The rest of this paper is arranged as follows: "\nameref{sec:rw}" provides the literature review, "\nameref{sec:method}" explains the word embedding technique, followed by the detailed explanation of deep convolutional forest (DCF), "\nameref{sec:result}" shows the results of the proposed method along with results of traditional machine learning classifiers and existing deep learning methods, "\nameref{sec:discussion}" discusses the findings and explains why the proposed solution outperforms the existing solutions. Finally, "\nameref{sec:conclusion}" concludes the proposed work and contains recommendations for future research.

\section*{Related work}
\label{sec:rw}

Over recent years, computer scientists have published a considerable volume of literature on spam detection \cite{Chan2015,Li2017,Akinyelu2021,Rao2021}; these works were limited to using machine learning and deep learning based models.

Bassiouni et al. \cite{Bassiouni2018} experimented multiple classifiers to filter emails gathered from the Spambase UCI dataset, which contained 4601 instances. They performed data preprocessing, then they selected features using Infinite Latent Feature Selection (ILFS). Finally, they classified emails with an accuracy of 95.45\% using Random Forest (RF), while the remaining classifiers: Artificial Neural Network (ANN), Logistic Regression, Support Vector Machine (SVM), Random Tree, K-Nearest Neighbors (KNN), Decision Table (DT), Bayes Net, Naive Bayes (NB) and Radial Basis Function (RBF) scored 92.4\%, 92.4\%, 91.8\%, 91.5\%, 90.7\%, 90.3\%, 89.8\%, 89.8\% * and 82.6\%, respectively.

Merugu et al. \cite{Merugu2019} classified text messages into Spam and Ham category with an accuracy rate of 97.6\% using Naive Bayes, which proved to outperform other machine learning algorithms such as Random Forest, Support Vector Machine and K-Nearest Neighbors according to the experimental results. The messages were collected from the UCI repository, which contained 5574 variable-length messages. To feed data into a classification model, the authors converted messages into fixed-length numerical vectors by creating term frequency-inverse document frequency (TF-IDF) {\cite{KIM201915}} vectors using the bag of words (BoW) model.

In 2020, Gaurav et al. \cite{Gaurav2020} proposed spam mail detection (SPMD) method based on the document labeling concept, which sorts the new messages into two categories: Ham and Spam. Experimental results illustrated that Random Forest produced the highest accuracy of 92.97\% among the three classification models: Naive Bayes, Decision Tree and Random Forest.

Lately, researchers have proposed deep learning methods such as Convolutional Neural Network (CNN) \cite{Popovac2018} and Short-Term Memory (LSTM) \cite{Barushka2018,Jain2019} for categorizing Spam and Ham messages. Popovac et al. \cite{Popovac2018} applied a CNN-based architecture after performing data preprocessing steps including tokenization, stemming, preservation of sentiment of text and removal of stop words. The feature extraction process involved transforming a text into a matrix of TF-IDF {\cite{KIM201915}} features. According to their experiment, CNN proved to be effective more than machine learning algorithms with an accuracy score of 98.4\%.

Another spam filtering model was proposed by \cite{Barushka2018}; they combined an N-gram TF-IDF feature selection, modified distribution-based balancing algorithm and a regularized deep multi-layer perceptron neural network model with rectified linear units (DBB-RDNN-ReL). Although their model was computationally intensive, the model provided an accuracy of 98.51\%.

Jain et al. \cite{Jain2019} used sequential stacked CNN-LSTM model (SSCL) to classify SMS spam with an accuracy of 99.01\%. They converted each text into semantic word vectors with the help of Word2vec, WordNet and ConceptNet. However, searching for the word vectors in these embeddings caused system overload.

Ghourabi et al. \cite{Ghourabi2020} presented a hybrid model for classifying text messages written in Arabic or English that is based on the combination of two deep neural network models: CNN and LSTM. The results indicated that the CNN-LSTM model scored an accuracy of 98.37\%, which is higher than other techniques like Support Vector Machine, K-Nearest Neighbors, Multinomial Naive Bayes, Decision Tree, Logistic Regression, Random Forest, AdaBoost, Bagging classifier and Extra Trees.

In 2020, Roy et al. \cite{Roy2020} focused on how to effectively filter Spam and Not-Spam text message downloaded from the UCI Repository \cite{Almeida2016}, which contains 5574 instances. They tested deep learning algorithms such as Convolutional Neural Network (CNN) and Long Short-Term Memory (LSTM) as well as machine learning classifiers such as Naive Bayes (NB), Random Forest (RF), Gradient Boosting (GB), Logistic Regression (LR) and Stochastic Gradient Descent (SGD). The experimental results confirmed that applying CNN with three convolutional layers and a regularization parameter (dropout) on randomly sampled 10-fold cross validation data resulted in an accuracy of 99.44\%. However, the authors spent much effort in tuning hyper-parameters.

As mentioned in "\nameref{sec:intro}", this study aims to handle feature relationships in textual data by using convolutional layers together with pooling layers as alternatives to the multi-grained scanning procedure proposed by \cite{Zhou2019}. Zhou and Feng \cite{Zhou2019} proposed an ensemble of ensembles mechanism, meaning that each level learns from its previous level and each level has an ensemble of decision-tree forests.

Descriptions of existing text-based spam detection techniques with respect to datasets, feature extraction and selection methods, types of algorithms, and performance measure are covered in Appendix {\ref{sec:appendix}}.

\section*{Methodology}
\label{sec:method}

\begin{figure*}[ht]
\centering
\includegraphics[width=\textwidth]{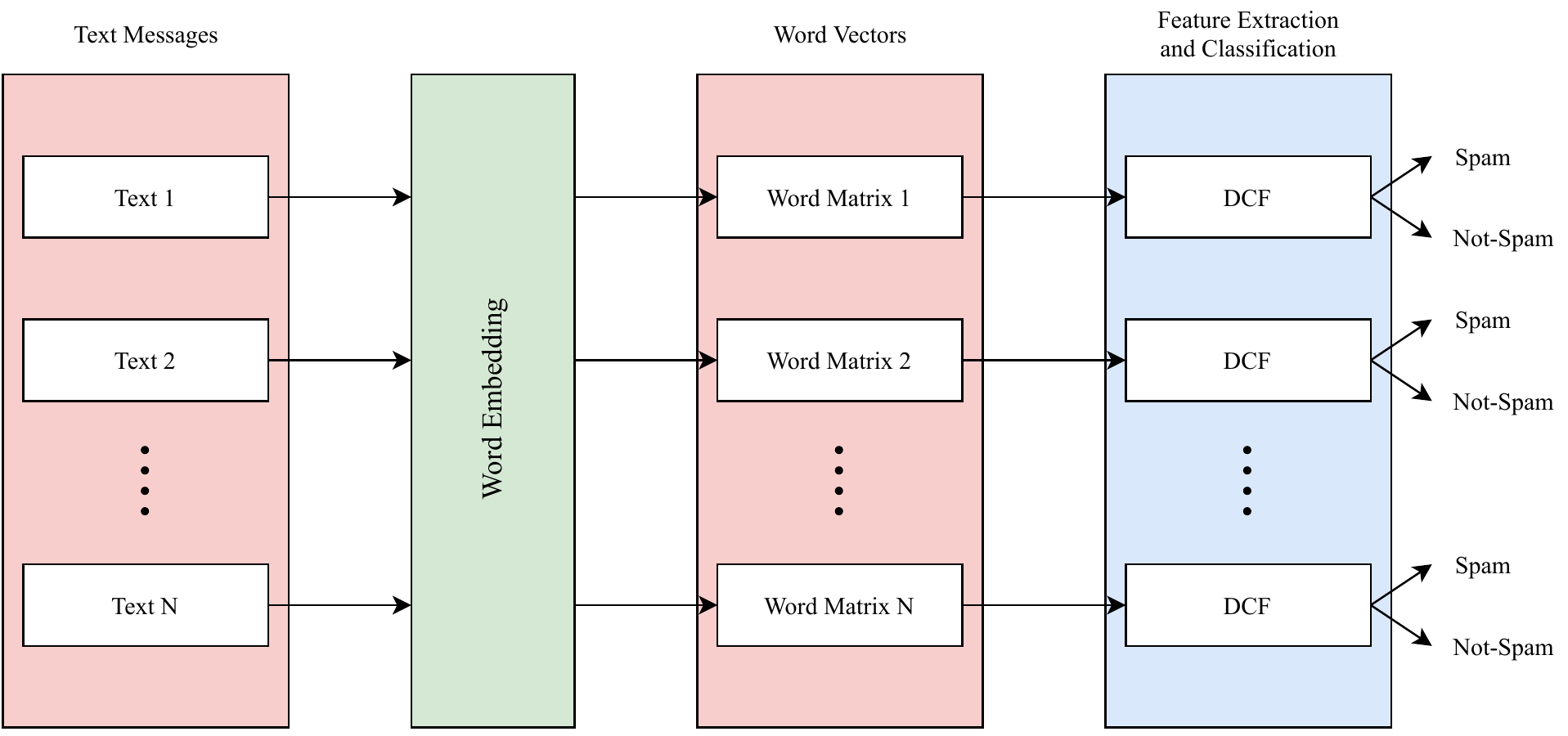}
\caption{Workflow of data: text preprocessing, word embedding, feature extraction and classification.}
\label{fig:Workflow}
\end{figure*}

\begin{figure*}[ht]
\centering
\includegraphics[width=\textwidth]{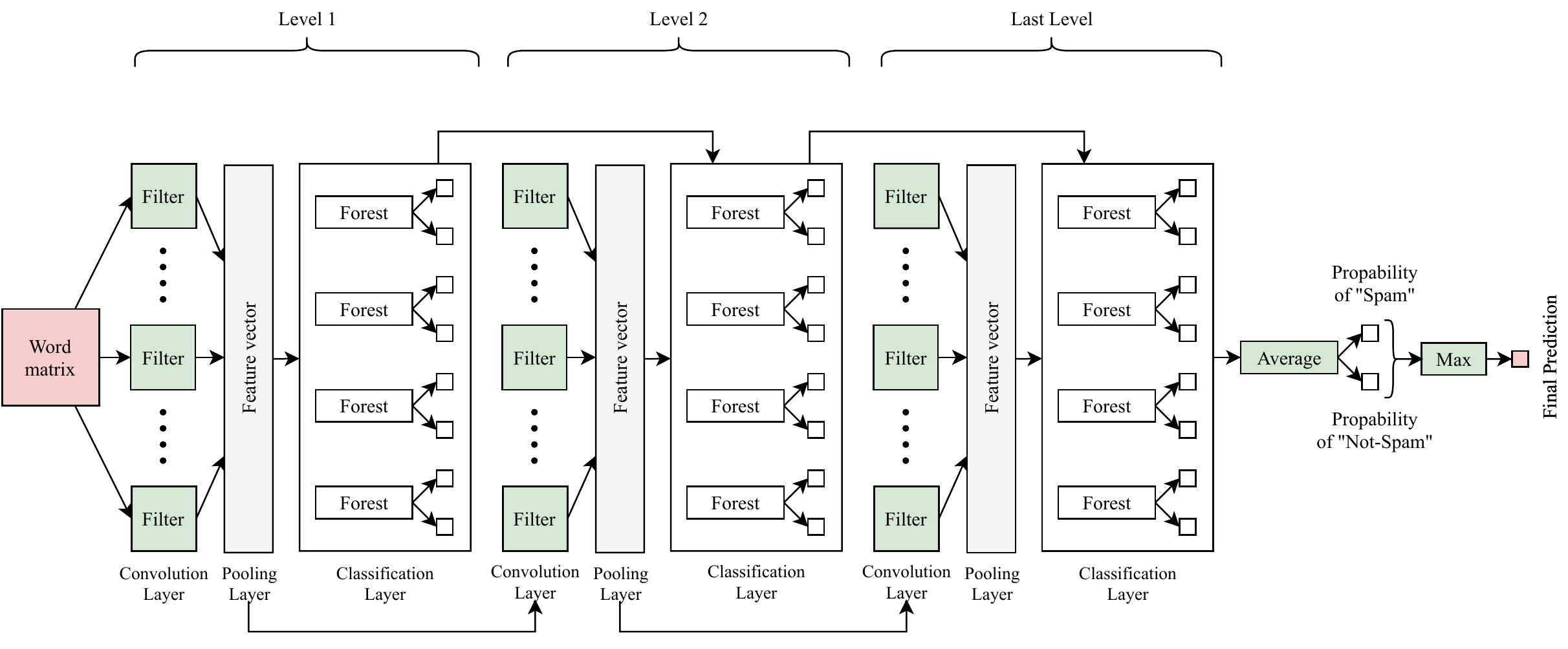}
\caption{Structure of deep convolutional forest (DCF).}
\label{fig:DCF structure}
\end{figure*}

Data pass through two main phases as shown in Figure \ref{fig:Workflow}: the first phase is applying the word embedding technique after preprocessing to convert textual data into a numerical form and the second phase is using deep convolutional forest (DCF) to extract features and classify text as illustrated in Figure \ref{fig:DCF structure}. The proposed method analyses the SMS spam dataset, which is publicly available \cite{Almeida2016}. The dataset has a collection of messages where each message is either 1 (Spam) or 0 (Not-Spam). First of all, text messages were prepared by splitting each message into a list of words and then performing text preprocessing techniques like stemming and stop words removal. Afterward, each word was transformed into a sequence of numbers called a word vector; the word vector is generated using a word embedding technique as explained in "\nameref{sec:wordembedding}". Finally, the generated word vectors are sent as a word matrix to DCF for classification and determining whether a message is Spam or Not-Spam as explained in "\nameref{sec:DCF}".

\subsection*{Word embedding}
\label{sec:wordembedding}

Word embedding is a technique where each word is represented by a vector holding numbers indicating the semantic similarity to other words in text corpus (i.e., similar words have similar representations). The difference between the word embedding and the one-hot encoding method is that the one-hot encoding method splits each text into a group of words and turns each word into a sequence of numbers disregarding the word meaning within context \cite{Elakkiya2020}, unlike the word embedding technique that transforms each word into a dense vector called a word vector that captures its relative meaning within the document \cite{Baccouche2020} using GloVe algorithm \cite{pennington2014glove} implemented by the embedding layer \cite{Mohamed2021}.

In word embedding technique, every message $m$ is a sequence of words: $w_1$, $w_2$, $w_3$, \ldots, $w_n$; each word is presented as a word vector of length $d$. After that, all word vectors of a given message (i.e., $n$ word vectors) are concatenated to form a word matrix $\textbf{M}\in\mathbb{R}^{n \times d}$. Finally, DCF receives the word matrix through the input layer and performs the convolution operation to produce feature maps through the convolutional layer.

\subsection*{Deep convolutional forest}
\label{sec:DCF}

The deep convolutional forest (DCF) model employs a cascade approach inspired by the structure of deep neural networks and deep forest \cite{Zhou2019} as shown in Figure \ref{fig:DCF structure}. Each level receives processed feature information from its prior level and outputs its processing outcome to the posterior level. The output of each level is the probabilities of both classes: Spam and Not-Spam; these probabilities are then concatenated with the feature maps to form the input of the next level. The model predicts the class of a given message by taking the average of the probabilities of Spam and the probabilities of Not-Spam separately from the last level output, and then it takes the maximum average as a final prediction result.

The model accuracy is the main factor that determines the number of levels. Whenever the accuracy of validation data increases, DCF continues to generate new levels and stops when there is no significant improvement in the accuracy score, unlike deep neural networks where the number of hidden layers is a pre-defined parameter. As a result, DCF is applicable to different scales of datasets, not limited to large-scale ones, as it automatically adjusts its complexity by terminating the training process when adequate.

The core units of each level in DCF are the convolutional layer, the pooling layer and the classification layer, which contains four base classifiers: two random forests \cite{Breiman2001} and two extremely randomized trees \cite{Geurts2006}. The convolutional layer is responsible for the feature extraction task, whereas the pooling layer helps reduce overfitting in the proposed model. Moreover, the classification layer predicts the probabilities of Spam and Not-Spam for a given message.

DCF combines the advantages of two techniques: bagging and boosting, they work interchangeably for decreasing variance and bias \cite{Barushka2019,Injadat2020}. Bagging refers to a group of weak learners that learns from each other independently in parallel and combines the outcomes to determine the model average \cite{Barushka2019}, whereas boosting is a group of weak learners that learns from each other in series where the next learner tries to improve the results of the previous learner \cite{Barushka2019}. DCF represents bagging through combining outputs from each forest in the classification layer as well as using Random Forest as a base classifier, which combines predictions from each decision tree and outputs the model average. Moreover, DCF supports boosting since it keeps adding levels where the next level tries to correct errors present in the previous level.

\subsubsection*{Convolution operation}

The convolutional layer extracts hidden features from the textual data by performing the convolution operation on a word matrix and applying the Rectified Linear Unit (ReLU) activation function \cite{Relu2021} on the output. Let $\textbf{M}\in\mathbb{R}^{n \times d}$ be the input word matrix having $n$ words and $d$-dimensional word vector, a filter $\textbf{F}\in\mathbb{R}^{d \times k}$ slides over the input, resulting in a feature vector $\textbf{O}$ of dimension $n-k+1$, also known as a feature map, where $k$ is the region size of the kernel. The process of finding the feature vector assuming that $k=2$ is shown in \eqref{eqn:convolution}.

\medskip
Let
\begin{align*}
M &= 
\begin{bmatrix}
w_{11} & w_{12} & \ldots & w_{1d}\\
w_{21} & w_{22} & \ldots & w_{2d}\\
\vdots & \vdots & \vdots & \vdots\\
w_{n1} & w_{n2} & \ldots & w_{nd}\end{bmatrix}\,, \\
\shortintertext{and}
F &= 
\begin{bmatrix}
f_{11} & f_{21}\\
f_{12} & f_{22}\\
\vdots & \vdots\\
f_{1d} & f_{2d}
\end{bmatrix}\,.
\end{align*}
Then
\begin{equation}
M\odot F= O\rightarrow \begin{bmatrix}
O_{1}\\
O_{2}\\
\vdots\\
O_{n-2+1}
\end{bmatrix}
\label{eqn:convolution}    
\end{equation}

Where $\odot$ is the convolution operator, which results in a feature vector $\textbf{O}$ of length $(n-2+1)$ in which each feature is calculated as follows:

$O_1=w_{11}f_{11}+w_{12}f_{12}+\ldots+w_{1d}f_{1d}+w_{21}f_{21}+w_{22}f_{22}+\ldots+w_{2d}f_{2d}$

$O_2=w_{21}f_{11}+w_{22}f_{12}+\ldots+w_{2d}f_{1d}+w_{31}f_{21}+w_{32}f_{22}+\ldots+w_{3d}f_{2d}$

$O_i=w_{(n-1)1}f_{11}+w_{(n-1)2}f_{12}+\ldots+w_{(n-1)d}f_{1d}+w_{n1}f_{21}+w_{n2}f_{22}+\ldots+w_{nd}f_{2d}$

The feature vector $\textbf{O}$ passes through the Rectified Linear Unit (ReLU) activation function. As shown in \eqref{eqn:relu}, ReLU takes each value $O_i$ and returns $O_i$ if the value is positive; otherwise, it returns zero, meaning that ReLU finds the maximum value between $O_i$ and zero. This value is noted by $\hat{O_i}$. 

The output of the convolutional layer after applying several filters becomes a set of feature maps as each filter produces one feature map $\hat{\textbf{O}}$ of length $(n-2+1)$ having positive values only.

\begin{equation}
    \hat{O_i} = max(0,O_i)
\label{eqn:relu}    
\end{equation}

\subsubsection*{Pooling operation}

The pooling layer in DCF applies the pooling operation for downsampling feature maps to avoid overfitting \cite{Akhtar2020}. Pooling is a process that aggregates the output of each filter by pulling a small set of features out of large sets to knock down the amount of computation required for processing the next level. Hence, pooling should reduce overfitting, which arises from high model complexity and causes misclassification of unseen data as the model learns the noise in textual data \cite{Jain2019,Ghourabi2020,Elakkiya2020}. In addition, DCF supports the early stopping procedure, meaning that the model stops training when the performance starts to degrade. Pooling has three common variations \cite{Roy2020}; max-pooling is the one that yielded better results than min-pooling and average pooling. The max-pooling operation highlights the most present feature in each feature map by calculating the maximum value. The pooling size was configured to be equal to the input size, so the output vector $\widetilde{\textbf{O}}=[\widetilde{O_1},\widetilde{O_2},\ldots,\widetilde{O_L}]$ has $L$ features, where each element $\widetilde{O_i}$ is the maximum value of each feature map and can be calculated by \eqref{eqn:maxpooling}.

\begin{equation}
    \widetilde{O_i} = max(\hat{O})
\label{eqn:maxpooling}    
\end{equation}

In the end, the features extracted using the convolution and the pooling layer (i.e., $\widetilde{\textbf{O}}$) pass through the classification layer. Algorithm \ref{algorithm:feature extraction} provides the detailed steps of the feature extraction process, considering that the number of features is equal to the number of filters.

\begin{algorithm}[ht]
\caption{Feature extraction}
\label{algorithm:feature extraction}
\begin{algorithmic}[1]
\Require $M\in\mathbb{R}^{n \times d}$
\Ensure $\widetilde{O}$ of length $L$ (number of features)

\Statex
\Function{ExtractFeatures}{$M$}
    \For {$j = 1$ to $L$}
     \State $O_j \gets M \odot F_j$
     \State $\hat{O}_{j} \gets \Call{ReLU}{O_j}$
     \State $\widetilde{O}_j \gets \Call{Max}{\hat{O}_j}$
    \EndFor
    \State \Return $\widetilde{O}$
\EndFunction

\end{algorithmic} 
\end{algorithm}

\subsubsection*{Random forests and extremely randomized trees}

The base classifiers in the proposed model are random forests \cite{Breiman2001} and extremely randomized trees \cite{Geurts2006}, which rely on the decision tree ensemble approach. Ensemble methods improve prediction results by combining multiple classifiers for the same task \cite{zhou2012ensemble}. Each level in DCF involves different forest types to support diversity; diversity is crucial in constructing ensemble methods \cite{zhou2012ensemble}.

Both forest types: random forests \cite{Breiman2001} and extremely randomized trees \cite{Geurts2006}, consist of a vast number of decision trees, where the prediction of every tree participates in the final decision of a forest by taking the majority vote. Furthermore, the growing tree procedure is the same in both techniques as they select a subset of features randomly, but they have two exceptions, as explained below.

Random forests build a decision tree by subsampling the input and selecting a subset of features randomly. Then, choosing the optimum feature for the split at each tree node according to the one with the best Gini value \cite{Injadat2020}, whereas extremely randomized trees manipulate the whole input and choose a random feature for the split at each node.

\begin{algorithm}[ht]
\caption{Deep Convolutional Forest}
\label{algorithm:DCF}
\begin{algorithmic}[1]
\Require $M\in\mathbb{R}^{n \times d}$
\Ensure $Y$ (the message label: 0 or 1)

\Statex
\State $\widetilde{O} \gets \Call{ExtractFeatures}{M}$
\State $probabilities= \emptyset$
\Repeat
    \State $P_{ham}=\emptyset,P_{spam}=\emptyset$
    \For{$i = 1$ to $4$}
    	\State $P_h,P_s \gets \Call{Forest}{\widetilde{O}, probabilities}$
    	\State $P_{ham} \gets \Call{Concat}{P_h, P_{ham}}$
    	\State $P_{spam} \gets \Call{Concat}{P_s, P_{spam}}$
    \EndFor
    \State $probabilities \gets \Call{Concat}{P_{ham},P_{spam}}$
    \State $\widetilde{O} \gets \Call{UpdateFeatures}{\widetilde{O}}$
\Until no significant improvement in performance

\State $\bar{h} \gets \Call{Mean}{P_{ham}}$
\State$\bar{s} \gets \Call{Mean}{P_{spam}}$
\If{$\bar{h} > \bar{s}$}
  \State $Y=0$
\Else
  \State $Y=1$
\EndIf
\State \Return $Y$

\Statex
\Function{UpdateFeatures}{$X$}
    \State \textbf{initialize} $W \gets$ random weights
    \State $X \gets X \odot W$
    \State $\hat{X} \gets \Call{ReLU}{X}$
    \State $\widetilde{X} \gets \Call{Max}{\hat{X}}$
    \State \Return $\widetilde{X}$
\EndFunction
\end{algorithmic} 
\end{algorithm}

\subsubsection*{Final prediction}

As shown in Figure \ref{fig:DCF structure}, the DCF first level extracts features from a text through the convolutional layer and the max-pooling layer. Next, each forest in the first level classification layer takes the extracted features and outputs two probabilities: the probability of a given message being Spam and being Not-Spam. After that, DCF computes the accuracy of the first level to be compared with the new accuracy of the second level. The second level in DCF produces new features, which are then concatenated with the probabilities generated by the first level to form the input to the second level classification layer, which outputs new probabilities (predictions). A new accuracy is calculated and compared with the previous accuracy so that DCF will continue to generate levels until it finds no significant improvement in accuracy or reaches the maximum number of levels. Each level has a convolutional layer, a max-pooling layer and a classification layer consisting of four forests: two random forests \cite{Breiman2001} and two extremely randomized trees \cite{Geurts2006}. Hence, each level in DCF outputs eight probabilities (i.e., four probabilities for each class).

The last level in DCF takes the average of probabilities for Spam and the average of probabilities for Not-Spam; the higher average value will be the final prediction. Algorithm \ref{algorithm:DCF} provides the full implementation of DCF.

\section*{Experimental results}
\label{sec:result}

This section analyzes the model performance in detecting Spam messages gathered from the UCI repository \cite{Almeida2016} and compares the results with multi-grained cascade forest (gcForest) and the traditional machine learning classifiers as well as the existing deep learning techniques. As shown in Table \ref{table:SMS spam dataset}, the number of spam instances is extremely lower than the number of legitimate ones. Therefore, balancing the class distribution is necessary to obtain accurate results. Initially, the dataset was split into two subsets: 80\% of the messages are for training and the remaining 20\% are for testing and validation. Then, the SMOTE \cite{Chawla2002} over-sampling technique was applied for balancing data before feeding it into the classifier.

\begin{table}[htbp]
\centering
\caption{Statistics of the SMS spam dataset.}
\resizebox{\columnwidth}{!}{\begin{tabular}{l l l}
\hline
\textbf{Label} & \textbf{Number of instances} & \textbf{Class distribution (\%)}\\
\hline
Ham & 4825 & 86.59\\
Spam & 747 & 13.41\\
Total & 5572 & 100\\
\hline
\end{tabular}}
\label{table:SMS spam dataset}
\end{table}

\subsection*{Configuration}

The proposed approach was implemented with Python 3.7 along with TensorFlow \cite{tensorflow2015-whitepaper}, Keras \cite{chollet2015keras} and Scikit-learn \cite{scikitlearn}. The embedding layer in Keras converted textual data into word vectors using GloVe word embeddings, which contains pre-trained 100-dimensional word vectors. The convolutional layer yielded the most promising performance by using 64 filters for applying the convolution operation on each input, where each filter (kernel) is a two-dimensional array of weights that moves one unit at a time (i.e., stride is set to 1). Although many studies used a different number of filters in each convolutional layer in convolutional neural networks {\cite{Rao2021}}, DCF uses the same number of filters as there is no significant change in performance and to facilitate the process of tuning hyper-parameters. The experiment showed that max-pooling with a pool size equals the size of the input was better than min-pooling and average-pooling. Moreover, each forest in the classification layer contained 100 trees. However, using more trees failed to increase the accuracy. All the remaining parameters of the base classifiers were set to default. Consequently, the proposed model predicted Spam messages with an accuracy equals 98.38\% after generating two levels only. Table \ref{table:DCF configuration} summarizes the configuration setup of DCF.

\begin{table}[htbp]
\centering
\caption{Configuration setup of DCF.}
\resizebox{\columnwidth}{!}{\begin{tabular}{l l l}
\hline
\textbf{Parameter} & \textbf{Best value} & \textbf{Location}\\
\hline
Number of filters & 64 & Convolutional layers\\
Kernel size & 2 & Convolutional layers\\
Stride & 1 & Convolutional layers\\
Number of trees & 100 & Base classifiers\\
Number of generated levels & 2 & DCF\\
\hline
\end{tabular}}
\label{table:DCF configuration}
\end{table}

\subsection*{Evaluation metrics}

As discussed in "\nameref{sec:DCF}", the accuracy score is the main factor in determining the number of DCF levels. After each level, DCF estimates the performance on the validation set until it finds no significant gain in performance. The experiment showed that two levels were enough to classify Spam messages. As a result, the training procedure was terminated and the model was evaluated on the test set based on the following well-known classification metrics:

\begin{itemize}
\item[--] True Positives ($T_{P}$): number of positive (Spam) messages that are correctly predicted as positive (Spam).
\item[--] True Negatives ($T_{N}$): number of negative (Not-Spam) messages that are correctly predicted as negative (Not-Spam).
\item[--] False Positives ($F_{P}$): number of negative (Not-Spam) messages that are predicted as positive (Spam).
\item[--] False Negatives ($F_{N}$): number of positive (Spam) messages that are predicted as negative (Not-Spam).
\item[--] Precision: determines the ability not to label a negative (Not-Spam) message as positive (Spam).
    \begin{equation}
        Precision = \frac{T_{P}}{T_{P}+F_{P}}
    \label{eqn:precision}
    \end{equation}
\item[--] Recall: determines the ability to find all positive (Spam) messages.
    \begin{equation}
        Recall = \frac{T_{P}}{T_{P}+F_{N}}
    \label{eqn:recall}
    \end{equation}
\item[--] F1-score: is a weighted average of precision and recall.
    \begin{equation}
        F1{\text -}Score = 2\times \frac{Precision\times Recall}{Precision+Recall}
    \label{eqn:f1-score}
    \end{equation}
\item[--] Accuracy: compares the set of predicted labels to the corresponding set of actual labels.
    \begin{equation}
        Accuracy = \frac{T_{P}+T_{N}}{T_{P}+F_{P}+T_{N}+F_{N}}
    \label{eqn:accuracy}
    \end{equation}
\item[--] Receiver Operating Characteristic (ROC) curve: plots True Positive Rate (TPR) on y-axis as defined in \eqref{eqn:true positive rate} and False Positive Rate (FPR) on x-axis as defined in \eqref{eqn:false positive rate}. The area under the ROC curve (AUC) measures the model performance; the higher the AUC value, the better model.
    \begin{equation}
        TPR=\frac{T_{P}}{T_{P}+F_{N}}
    \label{eqn:true positive rate}    
    \end{equation}
    
    \begin{equation}
        FPR=\frac{F_{P}}{F_{P}+T_{N}}
    \label{eqn:false positive rate}    
    \end{equation}
\end{itemize}

According to the confusion matrix described in Table {\ref{table:confusion matrix}}, the proposed algorithm identified Spam messages with low false-negative and false-positive rates. Moreover, from the data in Table {\ref{table:classification report}}, it can be seen that DCF performed well on the test set, resulting in high precision, recall, f1-score, accuracy, and AUC score.

\begin{table}[htbp]
\centering
\caption{The confusion matrix of DCF.}
\resizebox{\columnwidth}{!}{\begin{tabular}{l|l|c|c|c}
\multicolumn{2}{c}{}&\multicolumn{2}{c}{Actual\%}&\\
\cline{3-4}
\multicolumn{2}{c|}{} & Spam & Not-Spam\\
\cline{2-4}
\multirow{2}{*}{Predicted\%} & Spam & 83.64 & 1.44\\
\cline{2-4}
& Not-Spam & 0.18 & 14.74\\
\cline{2-4}
\end{tabular}}
\label{table:confusion matrix}
\end{table}

The goal of constructing ensemble models is to minimize the generalization error. As long as the individual learners are diverse and independent, the prediction error of the ensemble model decreases {\cite{KOTU201919}}. DCF encourages diversity by employing different structures of forests as base classifiers. Table {\ref{table:diversity report}} shows that the results of DCF having four forests of the same type (i.e., four random forests) are indeed worse than having four forests with diverse building strategies as shown in Table {\ref{table:classification report}}. Hence, the diversity affects the performance of detecting Spam messages.

\begin{table}[htbp]
\centering
\caption{Classification results of DCF with two levels and diverse forest types.}
\resizebox{\columnwidth}{!}{\begin{tabular}{l l l l l l}
\hline
\textbf{Label} & \textbf{Precision} & \textbf{Recall} & \textbf{F1-score} & \textbf{Accuracy} & \textbf{AUC}\\
\hline
Spam & 0.9880 & 0.9111 & 0.9480 & \multirow{2}{*}{98.38\%} & \multirow{2}{*}{0.989}\\
Ham & 0.9831 & 0.9979 & 0.9904\\
\hline
\end{tabular}}
\label{table:classification report}
\end{table}

\begin{table}[htbp]
\centering
\caption{Classification results of DCF with two levels and the same type of forests.}
\resizebox{\columnwidth}{!}{\begin{tabular}{l l l l l l}
\hline
\textbf{Label} & \textbf{Precision} & \textbf{Recall} & \textbf{F1-score} & \textbf{Accuracy} & \textbf{AUC}\\
\hline
Spam & 0.9939 & 0.9000 & 0.9446 & \multirow{2}{*}{98.29\%} & \multirow{2}{*}{0.955}\\
Ham & 0.9811 &  0.9989 & 0.9899\\
\hline
\end{tabular}}
\label{table:diversity report}
\end{table}

The cross-entropy loss detects if the model suffers from overfitting as computed in \eqref{eqn:loss}, where $y \in $ \{0,1\} is the true label of a single sample and $p$ is the predicted probability. In a good fit model, the loss should keep decreasing until reaching a point of stability whenever the number of levels is increasing. When using pooling layers during the experiment, the cross-entropy loss decreased from 0.101 to 0.084, while removing pooling layers caused an increase in the loss from 0.131 to 0.182. Upon further analysis, adding pooling layers after convolutional layers enhances the learning performance.

\begin{equation}
    L_{\log}(y,p) = -{(y\log(p) + (1 - y)\log(1 - p))}
    \label{eqn:loss}
\end{equation}

\subsection*{Machine learning classifiers}

\begin{figure}[htbp]
\centering
\includegraphics[width=\columnwidth]{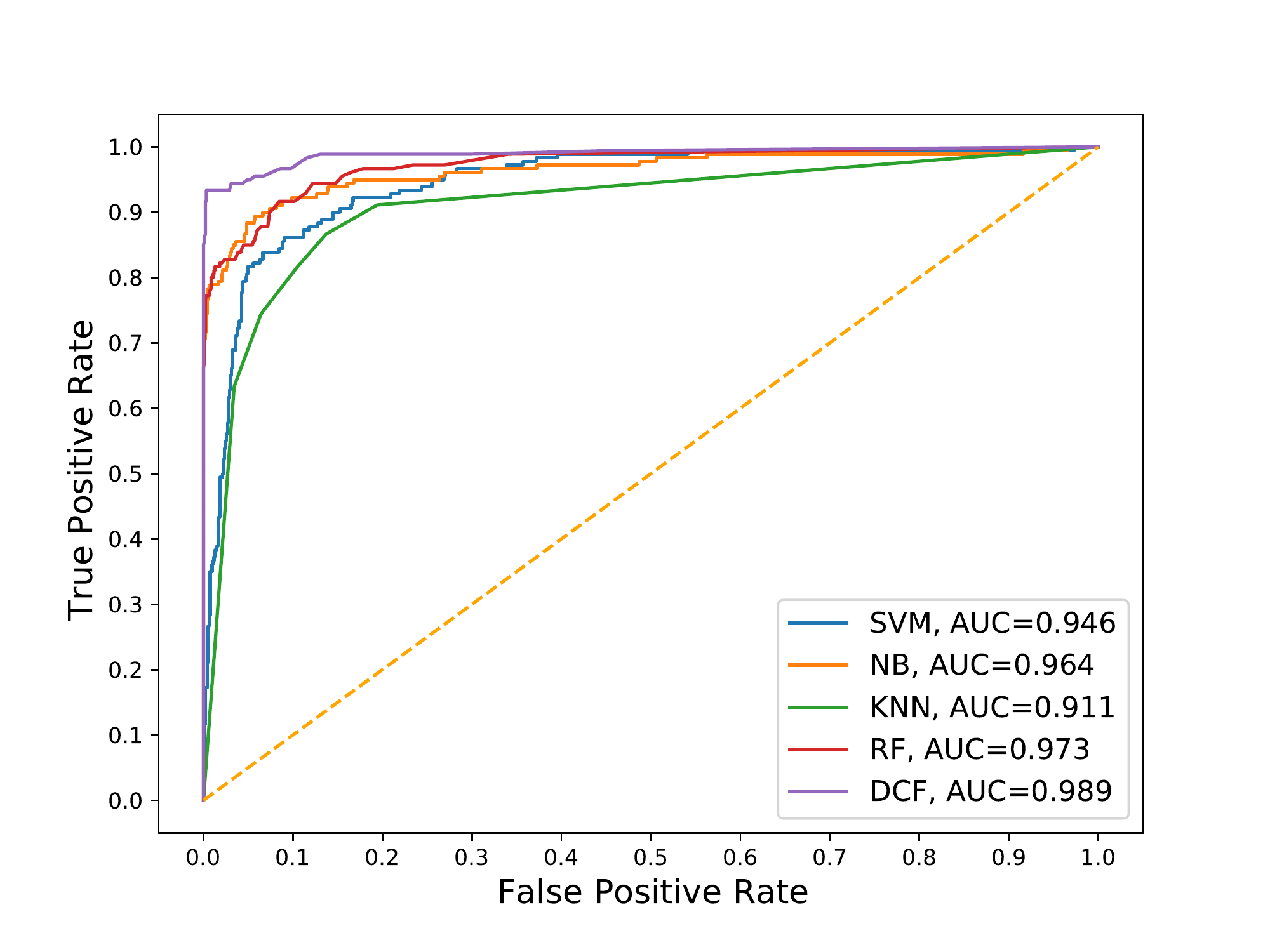}
\caption{ROC curve analysis of machine learning classifiers and the proposed model.}
\label{fig:ROC for ML models}
\end{figure}

To apply machine learning classifiers: Support Vector Machine (SVM), Naive Bayes (NB), K-Nearest Neighbors (KNN) and Random Forest (RF), text preprocessing techniques such as tokenization, removal of stop words and stemming were applied for extracting features manually from the SMS spam dataset. Table \ref{table:extracted features} presents 10 features that were extracted after data preprocessing. As indicated in Table \ref{table:classification report ML}, DCF outperformed other classifiers in categorizing Spam and Not-Spam messages in terms of precision, recall, f1-score and accuracy. According to the ROC curve in Figure \ref{fig:ROC for ML models}, the AUC score of the proposed model is significantly higher than the other classifiers, considering that the hyper-parameters of the mentioned machine learning algorithms were set to default during the experiment.

\begin{table*}[htbp]
\centering
\caption{The extracted features that are used in traditional machine learning classifiers}
\resizebox{\textwidth}{!}{\begin{tabular}{l l l}
\hline
\textbf{Feature} & \textbf{Description} & \textbf{Value} \\
\hline
characters count & number of characters (message length) & 1 or more \\
words count & number of words & 1 or more \\
readability score & Flesch–Kincaid readability test score, which indicates how difficult a text to understand \cite{Eleyan2020} & number \\
misspelled count & number of misspelled words & 0 or more \\
emails count & number of emails & 0 or more \\
phones count & number of phone numbers & 0 or more \\
is currency found & indicates the existence of currency symbol & 0 or 1 \\
IP address count & number of IP addresses & 0 or more \\
urls count & number of URLs & 0 or more \\
has blacklist url & indicates the existence of blacklisted URL & 0 or 1 \\
\hline
\end{tabular}}
\label{table:extracted features}
\end{table*}

\begin{table}[htbp]
\centering
\caption{Comparison of the results obtained by traditional machine learning classifiers and the proposed model.}
\resizebox{\columnwidth}{!}{\begin{tabular}{l l l l l l}
\hline
\textbf{Model} & \textbf{Label} & \textbf{Precision} & \textbf{Recall} & \textbf{F1-score} & \textbf{Accuracy}\\
\hline
\multirow{2}{*}{SVM} & Spam & 0.5142 & 0.9056 & 0.6559 & \multirow{2}{*}{84.64\%}\\
& Not-Spam & 0.9786 & 0.8349 & 0.9011 &\\
\hline
\multirow{2}{*}{NB} & Spam & 0.8938 & 0.7944 & 0.8412 & \multirow{2}{*}{95.15\%}\\
& Not-Spam & 0.9612 & 0.9818 & 0.9714 &\\
\hline
\multirow{2}{*}{KNN} & Spam & 0.6000 & 0.8167 & 0.6918 & \multirow{2}{*}{88.23\%}\\
& Not-Spam & 0.9620 & 0.8950 & 0.9273 &\\
\hline
\multirow{2}{*}{RF} & Spam & 0.9299 & 0.8111 & 0.8665 & \multirow{2}{*}{95.96\%}\\
& Not-Spam & 0.9644 & 0.9882 & 0.9762 &\\
\hline
\multirow{2}{*}{DCF} & Spam & \textbf{0.9880} & \textbf{0.9111} & \textbf{0.9480} & \multirow{2}{*}{\textbf{98.38\%}}\\
& Not-Spam & \textbf{0.9831} & \textbf{0.9979} & \textbf{0.9904} &\\
\hline
\end{tabular}}
\label{table:classification report ML}
\end{table}

\subsection*{Deep learning techniques}

Convolutional neural networks (CNN) and long short-term memory (LSTM) were implemented to compare their results with DCF. The number of convolutional layers affects the performance of CNN \cite{Roy2020}. Hence, three models of CNN were applied: the first model has one convolutional layer (1-CNN), the second model has two convolutional layers (2-CNN) and the third model has three convolutional layers (3-CNN). All of the mentioned deep learning models start with an embedding layer to generate 100-dimensional word vectors using GloVe; these word vectors are then used as inputs to the convolutional layer or the LSTM layer to produce feature maps. The convolutional layer in CNN has 64 filters of size 2 to match the configuration of DCF, and the number of units in LSTM was also set to 64. The models used the ReLU activation function as well as applying the Adam optimizer to reduce the error rate, in addition to adding a max-pooling layer in CNN models to avoid overfitting. Finally, the output layer used the Softmax activation function as defined in \eqref{eqn:softmax} to find the final decision, where $z$ is the input and $k = 2$ is the number of classes (i.e., Spam and Not-Spam).

\begin{equation}
    \sigma(z_i) = \frac{e^{z_{i}}}{\sum_{j=1}^K e^{z_{j}}} \ \ \ for\ i=1,2,\dots,K
\label{eqn:softmax}    
\end{equation}

\begin{table}[htbp]
\centering
\caption{Comparison of the results obtained by deep learning techniques and the proposed model.}
\resizebox{\columnwidth}{!}{\begin{tabular}{l l l l l l}
\hline
\textbf{Model} & \textbf{Label} & \textbf{Precision} & \textbf{Recall} & \textbf{F1-score} & \textbf{Accuracy}\\
\hline
\multirow{2}{*}{1-CNN} & Spam & 0.9588 & 0.9056 & 0.9314 & \multirow{2}{*}{97.84\%}\\
& Not-Spam & 0.9820 & 0.9925 & 0.9872 &\\
\hline
\multirow{2}{*}{2-CNN} & Spam & 0.9412 & 0.8889 & 0.9143 & \multirow{2}{*}{97.30\%}\\
& Not-Spam & 0.9788 & 0.9893 & 0.9840 &\\
\hline
\multirow{2}{*}{3-CNN} & Spam & 0.8907 & 0.9056 & 0.8981 & \multirow{2}{*}{96.68\%}\\
& Not-Spam & 0.9817 & 0.9786 & 0.9801 &\\
\hline
\multirow{2}{*}{LSTM} & Spam & 0.9477 & 0.9056 & 0.9261 & \multirow{2}{*}{97.66\%}\\
& Not-Spam & 0.9819 & 0.9904 & 0.9861 &\\
\hline
\multirow{2}{*}{DCF} & Spam & \textbf{0.9880} & \textbf{0.9111} & \textbf{0.9480} & \multirow{2}{*}{\textbf{98.38\%}}\\
& Not-Spam & \textbf{0.9831} & \textbf{0.9979} & \textbf{0.9904} &\\
\hline
\end{tabular}}
\label{table:classification report DL}
\end{table}

\begin{figure}[htbp]
\centering
\includegraphics[width=\columnwidth]{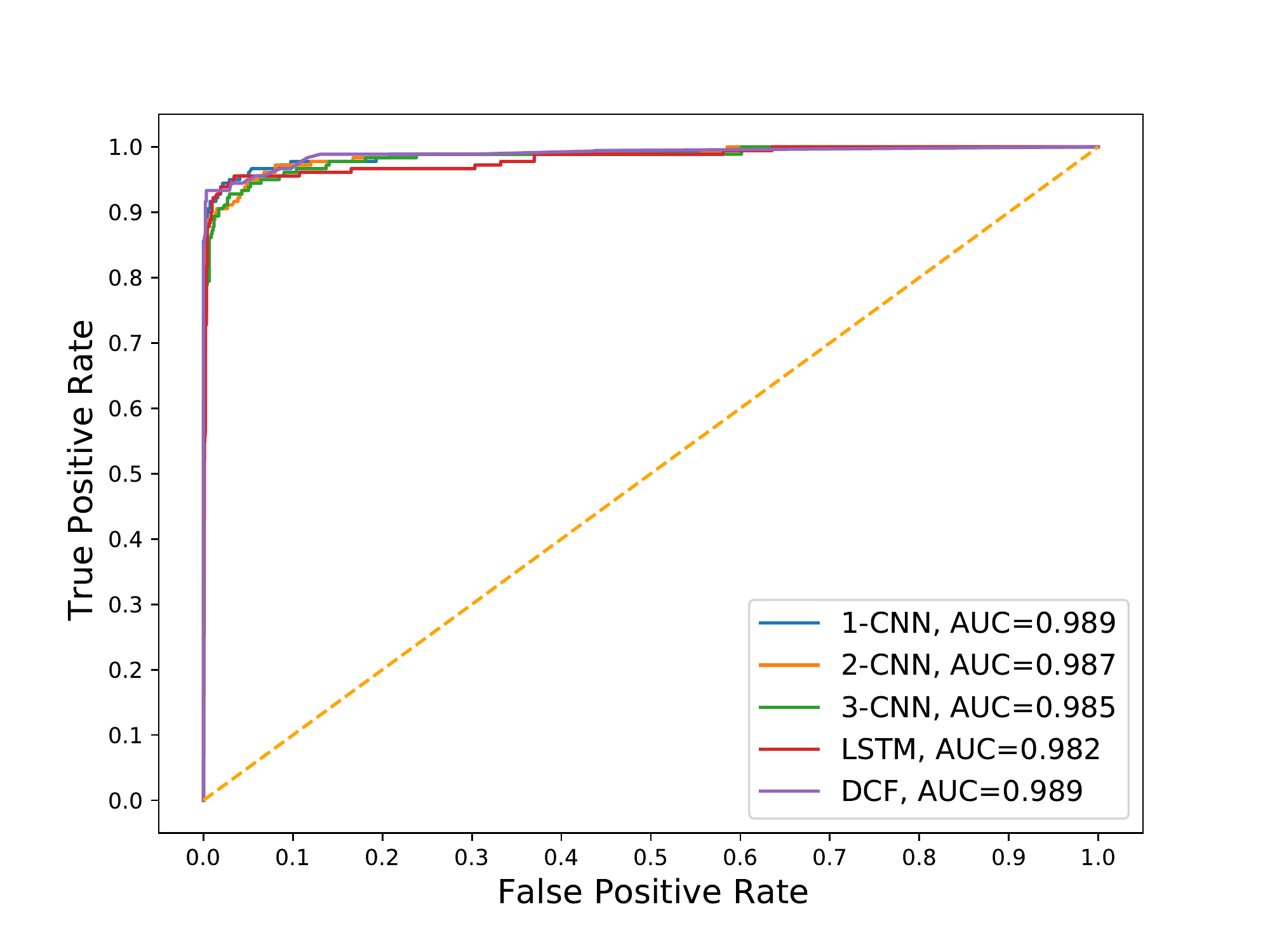}
\caption{ROC curve analysis of deep learning techniques and the proposed model.}
\label{fig:ROC for DL models}
\end{figure}

Table \ref{table:classification report DL} indicates that the proposed model realized the best performance with respect to precision, recall, f1-score and accuracy. In addition, DCF and 1-CNN achieved an equal AUC score, as indicated in Figure \ref{fig:ROC for DL models}.

\subsection*{Multi-grained cascade forest}
The gcForest model proposed in {\cite{Zhou2019}} depends on the multi-grained scanning procedure for feature extraction. However, this procedure is not capable of manipulating textual data. Therefore, in order to evaluate the gcForest model on the SMS spam dataset, the messages were represented by GloVe word embeddings before feeding to the model. Considering that the word embedding method conveys semantic relationships, unlike the TF-IDF method \cite{Dhanani2022}. Table {\ref{table:classification report gcForest}} signifies that DCF achieved better performance than gcForest. Nevertheless, TF-IDF features led to poor performance on DCF as the accuracy decreased to 75\% compared with word embeddings.

\begin{table}[htbp]
\centering
\caption{Comparison of the results obtained by multi-grained cascade forest (gcForest) and the proposed model.}
\resizebox{\columnwidth}{!}{\begin{tabular}{l l l l l l l}
\hline
\textbf{Model} & \textbf{Label} & \textbf{Precision} & \textbf{Recall} & \textbf{F1-score} & \textbf{Accuracy} & \textbf{AUC}\\
\hline
\multirow{2}{*}{gcForest} & Spam & 1.0000 & 0.7778 & 0.8750 & \multirow{2}{*}{96.40\%} & \multirow{2}{*}{0.983}\\
& Not-Spam & 0.9589 & 1.0000 & 0.9790\\
\hline
\multirow{2}{*}{DCF} & Spam & 0.9880 & \textbf{0.9111} & \textbf{0.9480} & \multirow{2}{*}{\textbf{98.38\%}} & \multirow{2}{*}{\textbf{0.989}}\\
& Not-Spam & \textbf{0.9831} & 0.9979 & \textbf{0.9904}\\
\hline
\end{tabular}}
\label{table:classification report gcForest}
\end{table}

\section*{Discussion}
\label{sec:discussion}

This paper introduced a dynamic (self-adaptive) deep ensemble technique to classify Spam and Not-Spam messages with remarkable classification results compared to the methods described in literature. The model suggested in this paper outperformed machine learning algorithms as well as deep learning models since ensemble learning connects the decisions from individual learners to improve the final decision. Moreover, DCF exceeded the outcomes of gcForest {\cite{Zhou2019}}, since DCF carries the high-level features of textual data and maintains the semantic relationships.

The introduced model extracted hidden features from data with the help of convolutional layers and pooling layers, unlike machine learning classifiers that require manual feature extraction from textual data, which requires domain knowledge. Furthermore, deep learning methods have fixed complexity, which means that they perform inefficiently on small-scale data. On the other hand, the proposed model can set the complexity automatically as the number of levels is determined according to the rate of accuracy increase, which means that it can perform efficiently on both small-scale data and large-scale data.

The main gaps in literature, which are addressed by the proposed algorithm, are stated as follows:
\begin{itemize}
\item[--] No domain expertise is required to carry out the classification process.
\item[--] Dynamic increase in the model complexity in proportion to the increase in performance.
\end{itemize}

To sum up, the model developed in this paper can separate legitimate text messages from fraudulent ones with high accuracy and low complexity. This filtering process will reduce the possibility of stealing people's sensitive data and will ensure that the users will be able to focus on messages from multiple industry sectors, which will help companies grow their businesses.

\section*{Conclusion}
\label{sec:conclusion}

This paper presents a dynamic deep ensemble model for categorizing text messages into Spam and Not-Spam. The model starts from passing the word embeddings through convolutional and pooling layers to dispense with manual feature extraction. Then, the model sends the feature maps to the classification layers where the base classifiers: two random forests and two extremely randomized trees carry out the predictions. Adopting ensemble techniques like boosting and bagging in constructing the model accomplished more accurate outcomes than single classifiers. Ensemble procedure is implemented by processing the input in a level-by-level manner until reaching the last level in which the average of class probabilities is calculated to take the highest average value representing the predicted label. This procedure facilitates the adjusting of the model complexity, unlike deep learning where the model complexity is determined in advance. As confirmed by the experimental findings, the proposed model surpassed the traditional machine learning classifiers as well as the existing deep neural networks in terms of precision, recall and f1-score in addition to achieving the highest accuracy rate of 98.38\%. Overall, the suggested solution in this paper can significantly minimize the risks related to security attacks such as SMS phishing by filtering spam messages. Future work may include the detection of spam content written in different languages other than English. Furthermore, a slight change in the model architecture may be considered for classifying messages that involve images.


\bibliographystyle{spmpsci}      
\bibliography{library.bib}   

\clearpage
\onecolumn
\begin{appendices}

\section{Existing text-based spam detection techniques}
\label{sec:appendix}

\begin{table*}[ht]
\centering
\resizebox{\textwidth}{!}{\begin{tabular}{R R R R R}
\hline

\textbf{Model} & \textbf{Dataset(s) used} & \textbf{Feature extraction/selection} & \textbf{Algorithm(s) used} & \textbf{Performance measure} \\
\hline

\cite{Bassiouni2018} & Spambase UCI	& ILFS	& RF with 100 trees, ANN, Logistic, SVM, Random Tree, KNN, DT, Bayes Net, NB and RBF	& The best accuracy = 95.45\% for RF \\

\cite{Merugu2019} & SMS spam & TF-IDF & NB, KNN, RF, SVM & The best accuracy is close to 98\% for NB, RF, SVM \\

\cite{Gaurav2020} & Enron, Ling-Spam, and PU & TF-IDF & SPMD with NB, DT and RF & The highest accuracy = 92.56\% for RF in Ling-Spam dataset \\

\cite{Popovac2018} & SMS spam & TF-IDF & CNN & 98.4\% \\

\cite{Barushka2018} & Enron, SpamAssassin, SMS, and Social networking & N-gram TF-IDF & DBB-RDNN-ReL & The highest accuracy = 99.79\% in SpamAssassin dataset \\

\cite{Jain2019} & SMS spam and Twitter & Word2Vec, WordNet and ConceptNet & SSCL & The highest accuracy = 99.01\% in SMS spam dataset \\

\cite{Ghourabi2020} & SMS spam and Arabic messages & Word2Vec and GloVe & CNN-LSTM & The highest accuracy = 98.37\% in SMS spam \\

\cite{Roy2020} & SMS spam & Word embeddings & CNN and LSTM & 99.44\% \\

\cite{Xia2021} & SMS spam & Word weighting & Hidden Markov Model (HMM) & 96.9\% \\

\cite{Liu2021} & SMS spam and Twitter & TF-IDF and GloVe & Spam Transformer & The highest accuracy = 98.92\% in SMS spam \\

\hline
\end{tabular}}
\end{table*}

\end{appendices}

\end{document}